\documentclass{article}
\usepackage{ijcai18}

\usepackage{times}
\usepackage{xcolor}
\usepackage{soul}
\usepackage[utf8]{inputenc}
\usepackage[small]{caption}

\usepackage{graphicx} 
\usepackage{subcaption}
\usepackage{multirow}
\usepackage{amsmath}
\usepackage{dsfont}
\usepackage{amssymb}

\newcommand{\beq}{\begin{equation}}
\newcommand{\eeq}{\end{equation}}

\usepackage{pifont}
\usepackage{makecell}
\usepackage{bm}
\newcommand*\samethanks[1][\value{footnote}]{\footnotemark[#1]}

\title{3D-PhysNet: Learning the Intuitive Physics of Non-Rigid Object Deformations}

\author{Zhihua Wang\thanks{These two authors contributed equally}
$^{1}$, Stefano Rosa\samethanks$^{1}$, Bo Yang$^{1}$, Sen Wang$^{2}$, Niki Trigoni$^{1}$ \and Andrew Markham$^{1}$ \\
$^{1}$Department of Computer Science, University of Oxford \\
$^{2}$School of Engineering and Physical Sciences, Heriot-Watt University~ \\
\{name.surname\}@cs.ox.ac.uk, s.wang@hw.ac.uk
} 

\begin{document}

\maketitle

\begin{abstract}
The ability to interact and understand the environment is a fundamental prerequisite for a wide range of applications from robotics to augmented reality. In particular, predicting how deformable objects will react to applied forces in real time is a significant challenge. This is further confounded by the fact that shape information about encountered objects in the real world is often impaired by occlusions, noise and missing regions e.g. a robot manipulating an object will only be able to observe a partial view of the entire solid. In this work we present a framework, 3D-PhysNet, which is able to predict how a three-dimensional solid will deform under an applied force using intuitive physics modelling. 
In particular, we propose a new method to encode the physical properties of the material and the applied force, enabling generalisation over materials. The key is to combine deep variational autoencoders with adversarial training, conditioned on the applied force and the material properties.
We further propose a cascaded architecture that takes a single 2.5D depth view of the object and predicts its deformation. Training data is provided by a physics simulator. The network is fast enough to be used in real-time applications from partial views. Experimental results show the viability and the generalisation properties of the proposed architecture.

\end{abstract}


%
%
%
%
%

\section{Introduction}
The capacity to employ common-sense reasoning, analogous to human intuition, is necessary for unconstrained interaction with an arbitrary environment. In particular, the ability to understand the effect of applied forces on objects is vital for general purpose robotic applications.
Traditionally, mobile navigation approaches consider all objects as static, rigid obstacles. Similarly, in robotic grasping applications, objects are often considered rigid and non-deformable at the perception level, with the resultant deformations only taken into account at the later control stage. 
However in the real world, many objects are non-rigid and change in shape/size when subjected to external forces. The ability to infer likely deformations is of great use for predictive control. Forward simulators based on Finite Element Models (FEM) are typically used to compute body deformations. Although they are highly accurate, a full mesh representation of a solid is required, unable to work with incomplete depth views of an object, such as those available to a robot in the real world. Moreover, they are computationally costly, unable to be used in real-time. As an alternative, we explore conditional deep models to learn the underlying physics of deformation.

Latent space generative models like \emph{Generative
Adversarial Networks} (GANs) and \emph{Variational AutoEncoders}
(VAEs) learn a mapping from a latent encoding
space to a data space. The latent space learned
by these models is often organised in a near-linear fashion, so that neighbouring points in latent space map to similar points in data space. 
Generative networks have been applied with success to the problem of reconstructing 3D objects from partial views or synthesizing 3D objects \cite{yang17,HRSC}. 

\emph{Conditional Variational Autoencoders} (cVAEs) offer a natural way to encode the effects of physical properties and applied forces. 
On the other hand, a useful property of GANs is that the discriminator network
implicitly learns a rich, feature-level similarity metric.
VAE-GANs, first introduced in \cite{Larsen16}, combine the ability of VAEs to encode data into a latent space and the ability of GANs to produce sharper, high quality models.

We propose a novel deep network that combines a variational autoencoder and a discriminator, trained on synthetic data from an FEM-based physics simulator. Given a single depth image of the deformable object 
and conditioning input which includes the properties of the material, the strength of the force, and the location of the force, the network is able to output a predicted 3-D deformation of the solid. This prediction can then be used for tasks as diverse as robot manipulation and grasping, terrain deformation assessment, and in general for predicting the effect of forces on non-rigid objects in the context of end-to-end learning of intuitive physical models of the environment.

The main intuition is that it is possible for the network to learn the properties of the body of interest from physical quantities that describe the elasticity and compression properties of the material. 
This enables the network to generalise over a wide range of materials, relaxing the need for a large training set.

The material composition of an object can be estimated from RGB images using available approaches such as the recent Differential Angular Imaging for Material Recognition framework \cite{dain17}, which was trained on the GTOS (Ground Terrain in Outdoor Scenes) material reflectance database, containing 40 material classes.

The conditions can be entered as real values in an efficient way, enabling arbitrarily fine-grained values. 
A single prediction is more than three orders of magnitudes faster that an equivalent FEM simulation, at the cost of lower resolution; this makes the approach useful for online evaluation for time-sensitive tasks. 

We evaluate the proposed approach on both synthetic and real 
objects, and discuss the effect of different encodings of the conditions.

The main contributions of this work are:
\begin{itemize}
\item 3D-PhysNet, a novel architecture for learning non-rigid body deformations of 3D objects with a conditional VAE-GAN architecture
\item We also propose a Cascaded-3D-PhysNet, with a 3D reconstruction GAN followed by a cVAE-GAN, for predicting deformations from a single depth image
\item A natural way of conditioning the network on continuous, real valued physical quantities that reflect the real world, rather than discrete material classes
\item To our best knowledge, this is the first work on learning 3D deformations from arbitrarily rotated depth image views
\end{itemize}



\section{Related Work}
\label{sec:relatedwork}

\subsection{Intuitive Physics}
In \cite{frank14} the authors estimated the elasticity parameters of an object by interacting with it and by correlating applied forces and the resulting surface deformations. 

In \cite{wu2015galileo} the authors first proposed to use deep generative networks for learning intuitive physics from videos, such as the effect of gravity and friction on objects rolling down a slope. The key idea was based on inverting a physics engine to obtain model dynamics from observations. 
Deep networks have been subsequently used for predicting the stability of tower blocks \cite{lerer2016learning} and object dynamics \cite{mottaghi2016newtonian}. 

Predicting how actions affect the world is an open challenge. In \cite{finn2016unsupervised}
a deep model was trained in an unsupervised way to predict action-conditioned future video images of moving objects, using a technique called Convolutional Dynamic Neural Advection (CDNA) and action-conditioned LSTMs. 
Another approach to predicting future video snippets given conditions was proposed in \cite{Vondrick2016GeneratingVW}.
Applications of intuitive physics to robotics have been recently explored in
\cite{byravan2017se3}. The network predicts rigid body motions from 3-D point cloud inputs given a force vector applied to it, using a layer that encodes per-pixel SE(3) transformations.

Recently, \cite{vda17} decouples the future prediction problem by learning an abstract physical representation of the world using a perception network, and using the physical representation as input to a physics engine and and a rendering engine in order to generate visual data, that can be then matched to the visual input.

\subsection{Generative Networks}
Deep generative models such as GANs 
\cite{goodfellow2014generative} and VAEs 
\cite{kingma2013auto} have recently shown outstanding results in high-dimensional representations and generalisation ability.
GANs have been successfully applied in a number of different domains such as natural language understanding \cite{ijcai2017-576}, learning of latent spaces \cite{chen2016infogan} and 3-D reconstruction \cite{yang17}.

In the original GAN formulation the discriminator network is trained to classify real and fake examples. However, the loss function can be difficult to converge and training is often unstable.
WGAN \cite{Arjovsky2017WassersteinG} proposed to use Wasserstein distance with weight clipping for stabilizing training.
Recently, \cite{gulrajani17} proposed to penalize the norm of the discriminator gradient with respect to its input, further improving training stability.

In \emph{Conditional GANs} (cGANs) the generated output is conditioned on external conditional information.
cGANs addresses problems where the input-to-output mapping is of the type one-to-many.

\emph{Invertible conditional GANs} (IcGANs) \cite{perarnau2016invertible}
combine an autoencoder with a cGAN and have been shown to be able to learn a good latent representation of the inputs.
Conditional GANs have been recently used to learn mappings between input and output images
with both paired \cite{pix2pix2016} and unpaired \cite{zhu2017unpaired} images.

Combining variational autoencoders with adversarial training has also gained popularity \cite{Larsen16,
Bao2017CVAEGANFI}. The idea is to exploit the feature representation in
the GAN discriminator to complement the VAE
reconstruction loss.


\section{The Proposed Approach}
\label{sec:approach}


\subsection{Problem Formulation}
\label{sec:probform}
The problem of predicting 3D deformations from incomplete depth views 
can be split into two problems: learning a reconstruction $f_{rec}$ from an input depth view $I$ to a full 3D shape $X$, and learning a smooth mapping $f_{def}$ that maps a 3D shape $X$ into a deformed 3D shape $Y$, given a condition $y$.
In our case, $I$, $X$ and $Z$ are discretized into voxel grids of dimensions $64^3$, while the condition is a vector of $n$ real values, therefore the problem is defined as:
\beq
X=f_{rec}(I) \hspace{1cm} (X,I \in {Z_2^{64}}^3, Z_2={0,1} ),
\label{eq:3drecmapping}
\eeq
\beq
Y=f_{def}(X,y) \hspace{0.5cm} (X,Y \in {Z_2^{64}}^3, Z_2={0,1}, y \in \mathbb{R}^n).
\label{eq:3ddefmapping}
\eeq

\subsection{Main Framework}

\begin{figure}[h]
\centering
\includegraphics[width=0.95\columnwidth]{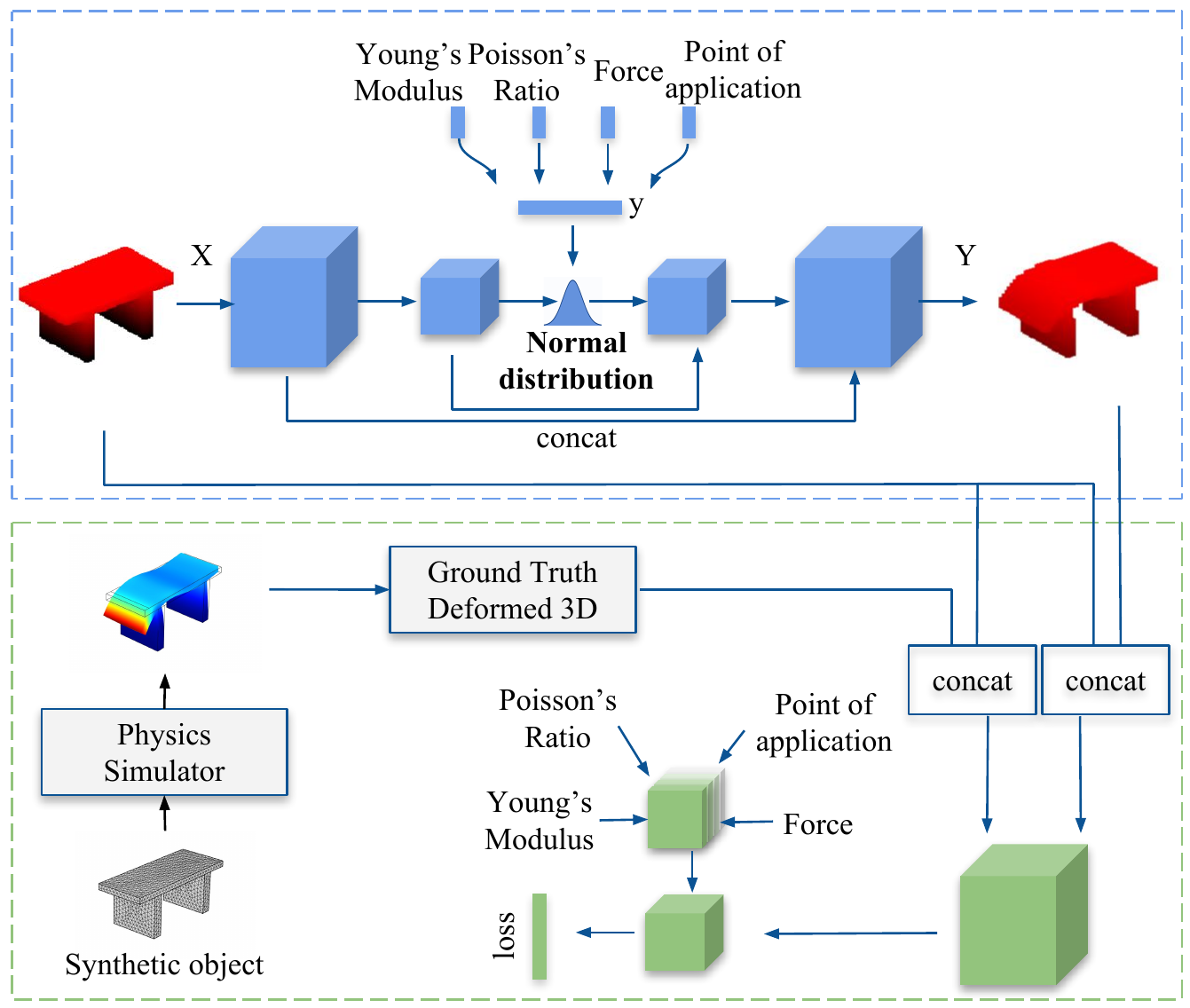}
\caption{The 3D-PhysNet architecture. Blue blocks represent the encoder E and generator G networks; green blocks represent the discriminator network D.}
\label{fig:model}
\end{figure}

\begin{table}[t]
\centering
\resizebox{0.8\columnwidth}{!}{
\begin{tabular}{l}
Encoder E  \\
\hline \\
$\lbrack$ input $\rbrack$ $64\times64\times64$      \\     
$\lbrack$ layer 1 $\rbrack$ Conv. $4^3$, Stride $2^3$, Max pool. $2^3$, ReLU activ.\\
$\cdots$      \\
$\lbrack$ layer 5 $\rbrack$ Conv. $4^3$, Stride $2^3$, Max pool. $2^3$, Sigmoid activ.\\
$\lbrack$ layer 6 $\rbrack$ FC 5000 Dense ReLU\\
$\lbrack$ layer 7 $\rbrack$ $\mu$ FC 800  $\sigma$ FC 800 \\

\\

Generator G \\
\hline \\
$\lbrack$ layer 1 $\rbrack$ FC 32768 Dense ReLU\\
$\lbrack$ layer 2 $\rbrack$ Deconv. $4^3$, Stride $2^3$, ReLU activ. CONCAT\\
$\cdots$      \\
$\lbrack$ layer 5 $\rbrack$ Deconv. $4^3$, Stride $2^3$, ReLU activ. CONCAT\\
$\lbrack$ layer 6 $\rbrack$ Deconv. $4^3$, Stride $2^3$, Sigmoid activ. CONCAT\\

\\

Discriminator D \\
\hline \\
$\lbrack$ input $\rbrack$ $64\times64\times64$ CONCAT      \\     
$\lbrack$ layer 1 $\rbrack$ Conv. $4^3$, Stride $2^3$, Max pool. $2^3$, ReLU activ.\\
$\lbrack$ layer 2 $\rbrack$ Conv. $4^3$, Stride $2^3$, Max pool. $2^3$, ReLU activ. \\CONCAT\\
$\cdots$      \\
$\lbrack$ layer 5 $\rbrack$ Conv. $4^3$, Stride $2^3$, Max pool. $2^3$, Sigmoid activ.\\
$\lbrack$ layer 6 $\rbrack$ FC 32768

\end{tabular}
}
\caption{Implementation details for the generator and the discriminator networks.}
\label{tab:networkparameters}
\end{table}




Figure \ref{fig:model} shows the architecture of the proposed 3D-PhysNet. It is composed of two main networks: a generator network $G$ and a discriminator network $D$, that are competing against each other. 

Broadly, the generator maps the undistorted 3-D model into a deformed 3-D model, conditioned on the supplied parameters. The discriminator is used during training only and is a classifier that determines whether its input is drawn from the ground-truth or the output of the generator. The generator and discriminator are adversarial i.e. they each get better over time. We now describe each network in detail.

\subsection{Generator}
The network takes as input a voxel grid of size 64$\times$64$\times$64, representing a 3-D point cloud. which is obtained by voxelizing the input 2.5D depth image.

The generator is implemented as a variational autoencoder network that is able to learn a normally distributed latent representation from  the input voxel grid.
To facilitate the replication of local structures and object details the convolutional and deconvolutional layers have skip connections between the input and the output samples.

The encoder E has five 3-D convolutional layers.
The encoder is followed by a fully-connected layer flattening the 3-D representation into a 1-dimensional vector, in turn followed by two layers $\mu$ and $\sigma$, representing the reparameterized mixture of gaussians from which we extract random samples.  

The condition vector encapsulates the material elasticity properties, the magnitude of the force, the location of the force (see Section \ref{sec:encoding}), and is concatenated with the latent vector.

The generator G follows the inverse of the encoder, composed of five deconvolutional layers which are followed by ReLU activations except for the last layer which is followed by a sigmoid function. 
The details of E and D are shown in Table \ref{tab:networkparameters}.

The loss $\mathcal{L}_{VAE}$ for the variational autoencoder is:
\beq
\mathcal{L}_{VAE} = \mathcal{L}_{ae} +  \mathcal{L}_{prior},
\eeq
where $\mathcal{L}_{ae}$ is a specialized form of Binary Cross-Entropy (BCE), as in \cite{brock2016generative}, and is given by:
\beq
  \mathcal{L}_{ae}  = -\alpha t \log(o) - (1-\alpha)(1-t) \log(1-o),
\eeq
where $t$ is the true binary value for each voxel ({0,1}), $o$ is the output value predicted by $E$ and is in the range (0,1), $\alpha$ is a parameter that balances false positives against false negatives. 

The second term of the loss is the Kullback-Leibler divergence of the latent representation from a normal distribution:
\beq
\mathcal{L}_{prior}  =  D_{KL}( N(\mu,\sigma) \vert\vert N(0,I) )
\eeq

The adversarial loss for the generator is described by:
\beq
\mathcal{L}^g_{gan} = -E \left[ D(o \vert x) \right]
\eeq
The total loss is:
\beq
\mathcal{L}^g = \beta \mathcal{L}_{VAE} + (1-\beta)\mathcal{L}^g_{gan},
\eeq
where $\beta$ is a weight that balances the VAE loss and the GAN loss.
Intuitively, the VAE loss guides the coarse 3D reconstruction of the object, and is important in the first phase of training, while the GAN loss is useful for learning to generate more plausible predictions, in particular the subtle shape deformations caused by the condition vector. 

\subsection{Discriminator}
The role of the discriminator is to evaluate whether the predicted deformations from the generator are realistic, by classifying them as real or fake compared to the real input. 

Similar to the encoder, it is composed of five 3-D convolutional layers. Implementation details reported in Table \ref{tab:networkparameters}.

The discriminator takes as input pairs of `real' ground truth voxel grids as informed by the physics simulator and 
`fake' generated voxel grids from the generator, in addition to the condition. The condition is reshaped so that it can be concatenated with the voxel grid. As the problem of shape deformation is high-dimensional, instead of outputting a binary value, the discriminator outputs a dense vector representing voxel similarities. 
The loss is based on WGAN-GP \cite{gulrajani17} which adopts Wasserstein Distance as a metric of similarity:

\begin{multline}
\mathcal{L}^d_{gan} = E \left[ D(o \vert x) \right] - E \left[ D(t \vert x) \right] + \\
\lambda E \left[ \left( {\lvert \lvert \nabla_{\hat{o}} D(\hat{o} \vert x) \rvert  \rvert }_2 -1 \right)^2 \right],
\label{eq:l_d_gan}
\end{multline} 
where $\hat{o} = \eta x + (1-\epsilon)o, \eta \in [0,1]$.

\subsection{Cascaded-3D-PhysNet}

\begin{figure}[t]
\centering
\includegraphics[width=0.9\columnwidth]{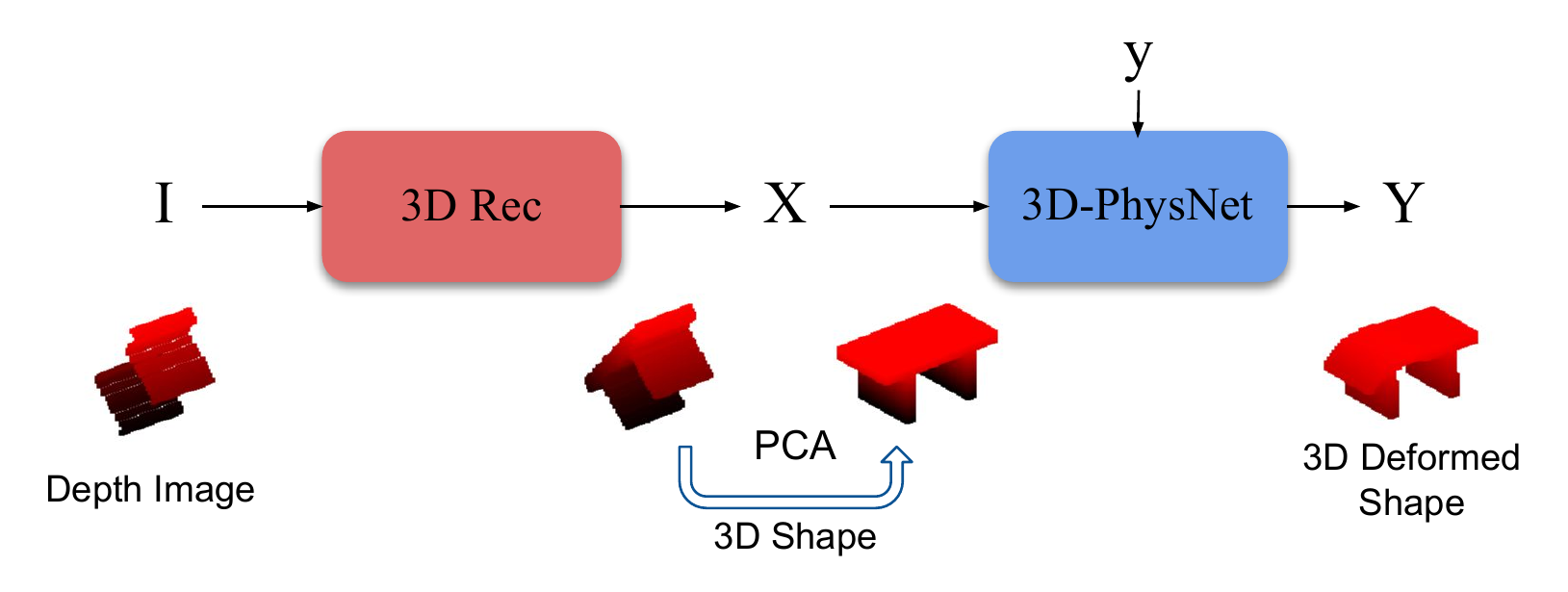}
\caption{Cascaded configuration with 3D reconstruction network (red) and 3D-PhysNet network (blue).}
\label{fig:cascadedmodel}
\end{figure}

We also propose a cascaded framework, Cascaded-3D-PhysNet, in which the output of a 3D reconstruction network is fed into the input of 3D-PhysNet, as shown in Figure \ref{fig:cascadedmodel}.
In this configuration we factor the learning of the mapping functions (Section \ref{sec:probform}) into two separate, independently trained problems, the first performing 3-D reconstruction from a 2.5D point cloud and the second responsible for deforming the 3-D model. In our experimental results section we demonstrate that this cascaded approach is superior to directly inputting 2.5D partial views into 3D-PhysNet and attempting to perform both reconstruction and deformation in a single network. Our intuition is that the latent encoding provided by the GAN is more suited for learning high resolution reconstruction, subject to arbitrary rotation, whereas the normal encoding of the VAE is better suited for representing the physically-based smooth functions representing the material properties and applied force. We are agnostic to the reconstruction network used, in our approach we adopt the recent 3D-RecGAN framework~\cite{yang17}. To simplify the task of 3D-PhysNet PCA is to used to align the rotated shape along its principal axes.

\subsection{Encoding Physical Properties} 
\label{sec:encoding}
In 3D-PhysNet, we take the novel approach of using the continuous material \textit{properties} as the conditioning input, rather than the material \textit{class}. This is more generalisable and can be arbitrarily quantized, enabling deformation estimation for previously unseen materials. 

$f_{def}$ describes the deformation of a solid, which can be defined as 
$$
f_{def}: \mathbf{x} \mapsto \mathbf{x} + \mathbf{d},
$$
where $\mathbf{x}$ is the set of points representing the undeformed solid and $\mathbf{d}$ represents a deformation field.
In our case we make the assumption that the body is composed of a homogeneous material, isotropic and linearly elastic. 

For continuous media the following relationship (an extension of the well-known Hooke's Law for uniaxial deformation) relates applied force (stress) to resultant deformation (strain):
$$
\bm{\sigma} = C \bm{\epsilon},
$$
where $\bm{\sigma}$ is the stress tensor, $\bm{\epsilon}$ is the strain tensor and $C$ is the Cauchy tensor mapping strain to stress.
Under the assumption that the material is isotropic, $C$ only depends on two physical parameters: the Young's modulus $E$ and Poisson's ratio $\nu$.

The Young's modulus describes the force needed to enlarge or compress a material by a fixed amount and is defined by the ratio of stress to strain in the direction of the applied force. In practice, $E$ denotes the stiffness (or its reciprocal, elasticity) of a material.

\begin{figure}[t]
\centering
\includegraphics[width=0.7\columnwidth]{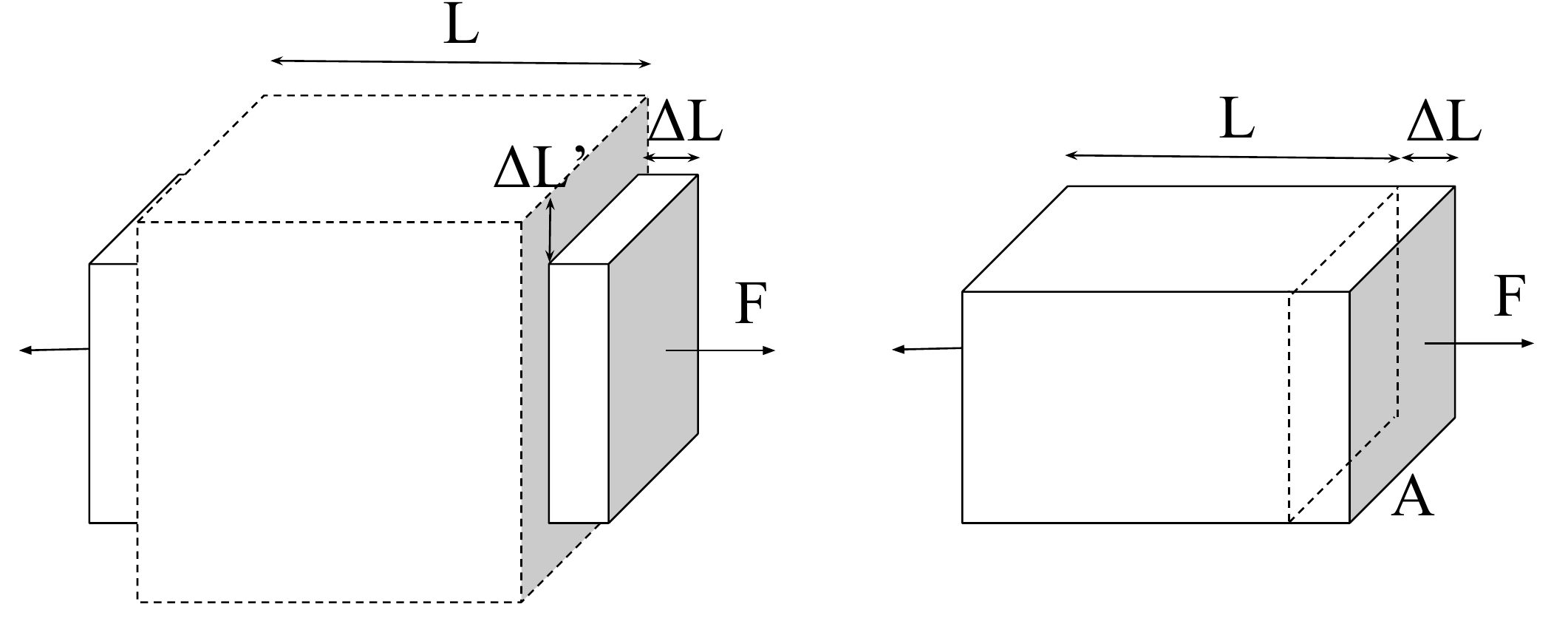}
\caption{Poisson's ratio $\nu$ (left) and Young's modulus $E$ (right) explained. $\nu$ shows how, given an applied force F which stretches the object along one axis, a resultant compression occurs along the other two axes. $E$ on the other hand represents the ratio of stress $\Delta L / L$ and strain $F / A$.}
\label{fig:elasticityparameters}
\end{figure}

The Poisson's ratio denotes the negative ratio of the transverse strain over the axial strain. When a material is compressed in one direction, an expansion is observed in the other two perpendicular dimensions, and vice versa.
In practice, $\nu$ denotes the compressibility of a material. 
Since we assume the material is isotropic, the Poisson's ratio is the same for every direction of compression/expansion (Figure \ref{fig:elasticityparameters}).

The choice of parameter sampling for the condition vector is critical and dependent on the problem of interest.
Young's modulus is measured in GigaPascals (GPa) in the SI system and is in the range $(0,\infty)$. For this reason, we fix an upper bound of $23$, which corresponds to the elasticity of concrete. 
Since $E$ varies over several orders of magnitude, we sample our training set logarithmically over the range and scale to $[0,1]$.  
The Poisson's ratio varies in the range $[0,0.5]$. Rubber has a Poisson's ratio of 0.5 (perfect volume conservation), while most materials are in the range 0.25 to 0.48. We therefore sample $\nu$ linearly over the full range.

\section{Experiments}
\label{sec:experiments}
\subsection{Physics Engine and Dataset Generation}
\label{subsec:physicsengine}


The physics engine is used to generate pairs of input and ground truth voxel grids and the corresponding condition vectors for training. 
In this work we used the COMSOL Multiphysics software in order to generate the training voxel grids, but the network is agnostic to the simulator.
The one-dimensional condition vector is obtained by concatenating the object's Young's modulus, the Poisson's ratio, the location of the force and its magnitude.


In order to generate our dataset for each object, we vary the Young's modulus and the Poisson's ratio over their whole range, using different sampling approaches, as discussed in the results. We vary the force magnitude over 30 values and the point of application of the force over 10 positions along the top of the object. We generate a second dataset in which we stretch the object along the x-y-z axes. For each axis we stretch over 8 scales. The finite element meshes used for the dataset generation contain on average 3200 triangles. We use a MUMPS solver on an Intel Xeon CPU with a corresponding simulation time for each sample of over 60~s.


We then extract a set of partial views for each sample. 
We create a virtual depth camera and rotate the object over a total of 125 different angles, with 5 uniformly sampled angles around
each rotation axis. For each rotation, a partial voxel grid is extracted representing a partial view.
The total number of samples is 50000. 
The full generated datasets will be released for the sake of reproducibility and testing.


\subsection{Implementation Details and Evaluation Setup}
The network was implemented with Tensorflow 1.4 and trained on a single Nvidia Pascal Titan X GPU. 
The network was trained with a batch size of 8 using the Adam optimizer, with $lr=5e-5$, 
$\beta 1=0.5$, $\beta 2=0.999$,$\epsilon =1e-8$.
$\lambda$ is set to 10 in Eq. \ref{eq:l_d_gan}.
The prediction time for a single input is 35.7ms, which is more than three orders of magnitude faster than an equivalent FEM simulation, allowing for real-time deformation estimation.

In the cascaded configuration we use 3D-RecGAN \cite{yang17} for shape reconstruction from rotated partial viewpoints. The network has a comparable prediction time.


We use voxel \emph{Intersection-Over-Union} (IOU) between two voxel grids as our accuracy metric. IOU is defined as:
$$
IOU = { { \sum_{ijk} \left[ \mathds{1}(Y_{ijk} > p) * \mathds{1}(X_{ijk}) \right]} 
\over { \sum_{ijk} \left[  \mathds{1}( \mathds{1}(Y_{ijk} > p) +  \mathds{1}(X_{ijk})) \right]} }
$$
where $\mathds{1}$ is an indicator function, $Y_{ijk}$ is the predicted voxel value at position $(i,j,k)$, $X_{ijk}$ is the true value of voxel $(i,j,k)$, and $p$ is the threshold for voxelization. In our experiments, $p$ is set as 0.8. 
The higher the IOU value, the better the reconstruction of a 3D model.




\subsection{Results and Discussion}
\label{sec:results}


\subsubsection{Latent Representation and Encoding of Physical Parameters}
In this experiment we compare the VAE latent encoding with a baseline inspired by IcGAN \cite{perarnau2016invertible}.
We also show the effects of the joint distribution over physical parameters on training time and convergence.
The convolutional layers of the GAN are the same as in our approach, while the latent vector in the GAN is a vector of dimension 5000. All other parameters, as well as the discriminator loss function, are the same as 3D-PhysNet.

\begin{figure}[t]
\centering
\includegraphics[width=\columnwidth,trim={0.5cm 0cm 1cm 0cm},clip]{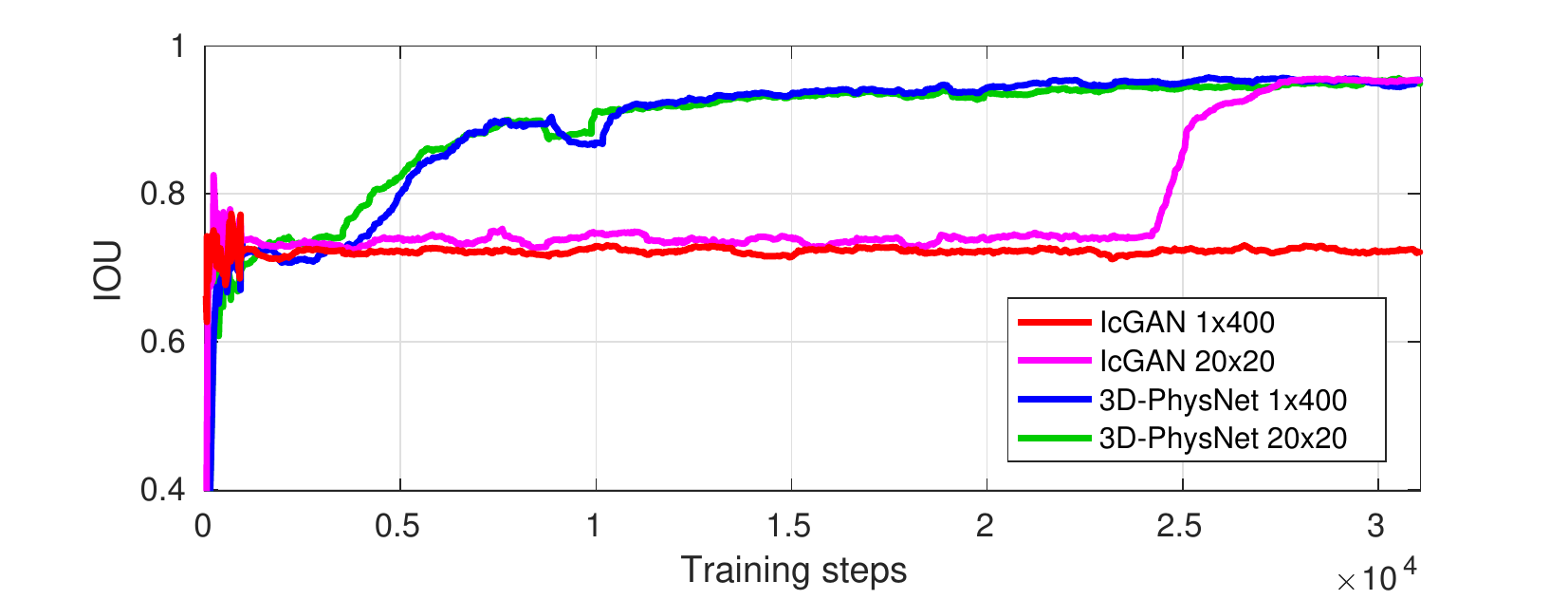}
\caption{Comparison with baseline and effects of the joint distribution over elasticity parameters.}
\label{fig:encodingnetworkcomparison}
\end{figure}

We simulate a bridge-like structure, and we first uniformly sample $E$ over 400 values keeping $\nu$ constant (termed 1$\times$400), then jointly vary both $E$ and $\nu$ over 20 values (termed 20$\times20$).  
Figure \ref{fig:encodingnetworkcomparison} shows the results of the four cases.

The baseline approach does not converge with the 1$\times$400 dataset, while it eventually converges with the 20$\times$20 one. The poor convergence can be intuitively explained by the fact that no implicit distribution is imposed on IcGAN and it needs to learn the mapping from each value to a particular deformation. For the case of sampling over 400 values of $E$ the dimensionality is too high for it to converge, whereas for 20$\times$20 the sparser sampling in each dimension allows it to converge after 25k iterations.

3D-PhysNet is able to converge quickly and predict deformation regardless of the dataset. This demonstrates that the normally distributed representation of the latent variables in the VAE is more suited to represent the smooth underlying function capturing the relationships between the conditions and deformation, compared with the arbitrary vector used in IcGAN. The learning curve for the  20$\times$20 dataset is marginally faster, possibly due to an underlying joint-relationship between $E$ and $\nu$ and the resultant Cauchy stress tensor.

\subsubsection{Generalisation over Physical Parameters and Scales}
In this experiment we analyse the ability of the network to generalise over the various conditions.

For this experiment we train 3D-PhysNet with full 3D inputs and vary both $E$ and $\nu$ over 20 values each. We then test on unseen values of $E$ and $\nu$. The network converges after 14k iterations and has a resultant IOU of 0.98, demonstrating good generalization to arbitrary materials.
In all our experiments the object is always resting on the ground.

\begin{figure}[b]
\centering
\includegraphics[width=0.75\columnwidth]{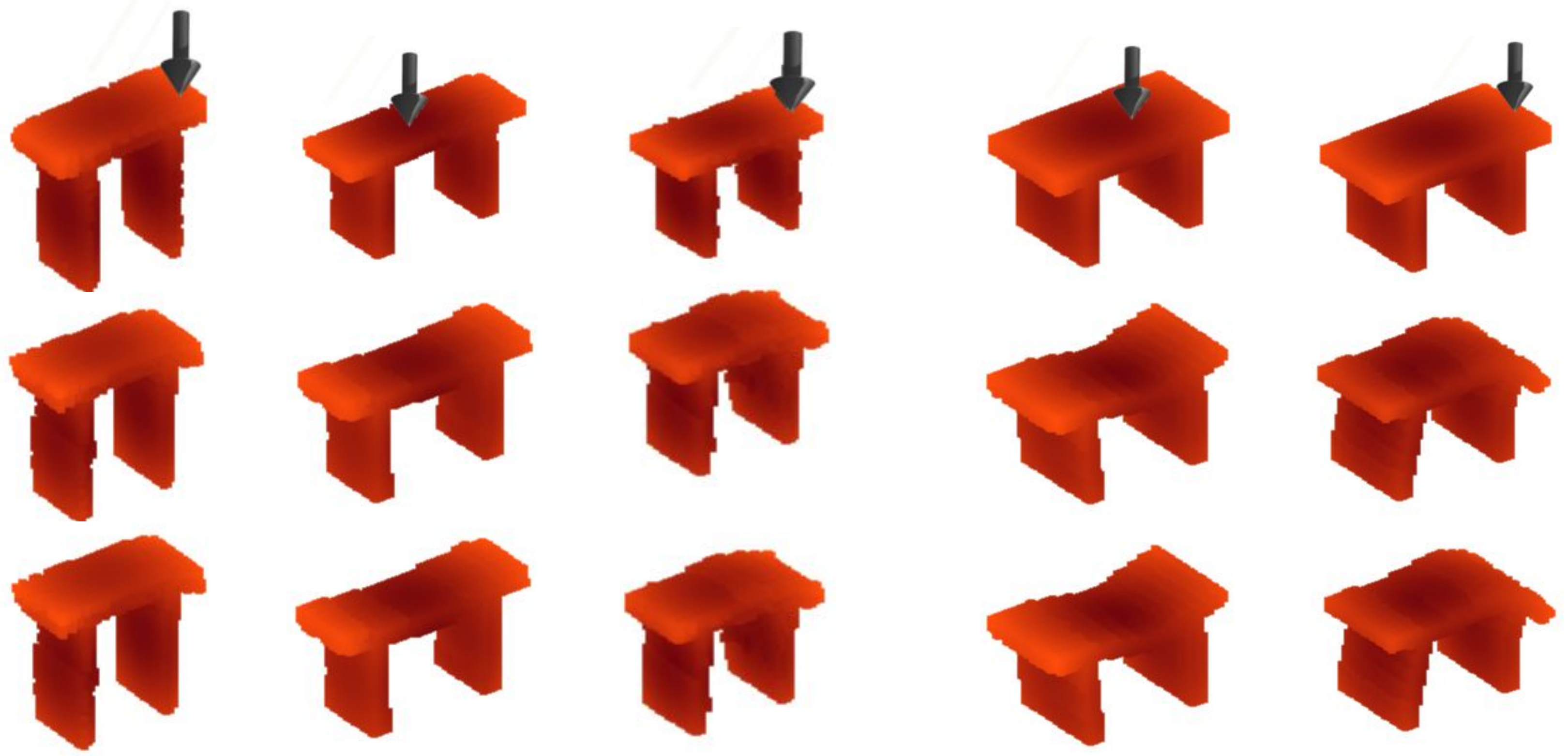}
\caption{Generalisation over shape variations and applied forces. Top row: undeformed object and applied force; middle: predicted deformations; bottom row: ground truth.}
\label{fig:shapes}
\end{figure}

We then evaluate the generalisation capabilities over scale variations. 
It should be noted that the scale variations affect the physics of the object, adding to the non-linearity of the problem.
The final average IOU for this experiment is 0.98; some qualitative results are shown in Figure \ref{fig:shapes}. 


We then investigate the ability of the network to generalise over force strength and location.
We train our network on a subset of the dataset composed by 310 samples: 31 different force strengths and 10 points of applications equally spaced over the top of the bridge. 
By encoding the force location as a real number the network only achieves a final average IOU of 0.9, while a one-hot encoding of the force location achieves an IOU of 0.95.
This shows that the force location is the most challenging parameter, since the resulting deformations for different positions of the force are widely different and not smooth. Qualitative results are shown in Figure \ref{fig:shapes}.

\subsubsection{Prediction from Partial Views and Cascaded Architecture}
In this experiment we motivate the use of a cascaded architecture for deformation prediction from partial views.

\begin{figure}[t]
\centering
\includegraphics[width=\columnwidth,trim={0.5cm 0cm 1cm 0cm},clip]{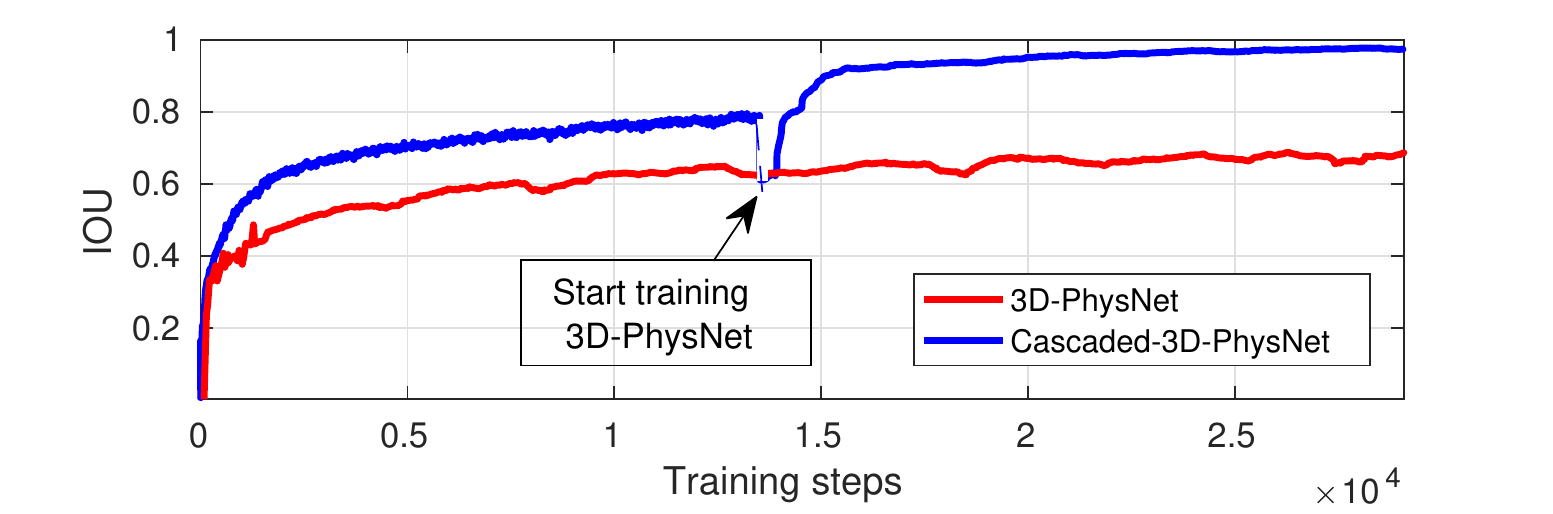}
\caption{Learning curves from partial views.}
\label{fig:rotatedviewpoints}
\end{figure}

In Figure \ref{fig:rotatedviewpoints} we show the learning curves for 3D-PhysNet and Cascaded-3D-PhysNet on the full dataset of rotated partial views.
For Cascaded-3D-PhysNet, we first train 3D-RecGAN on 3D reconstruction only, then 3D-PhysNet on the PCA-aligned outputs of 3D-RecGAN. We plot the two learning curves consecutively for comparison with the other approaches. 
It can be seen how 3D-RecGAN learns to reconstruct the objects fast, enabling 3D-PhysNet to converge faster.

\begin{figure}[b]
\captionsetup[subfigure]{justification=centering}
    \begin{subfigure}[t]{0.24\columnwidth}
        \centering
        \includegraphics[width=2cm,trim={1.5cm 1.5cm 1.5cm 1.5cm},clip]{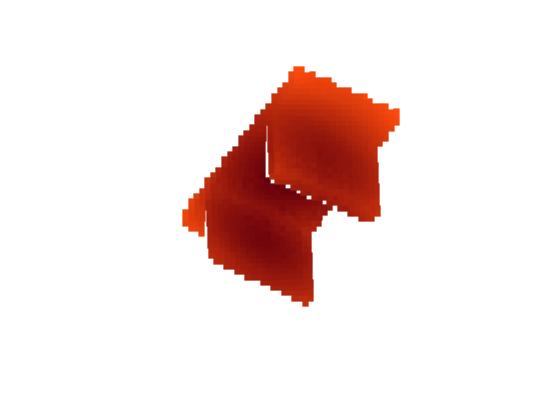} \\
        \vspace{-0.2cm}
        \includegraphics[width=2cm,trim={4cm 4cm 4cm 4cm},clip]{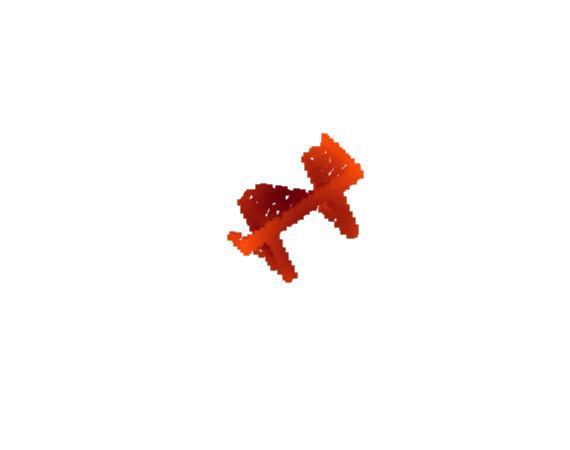}
        \caption{Input view}
    \end{subfigure}%
    ~ 
    \begin{subfigure}[t]{0.24\columnwidth}
        \centering
                \includegraphics[width=2cm,trim={4cm 4cm 4cm 4cm},clip]{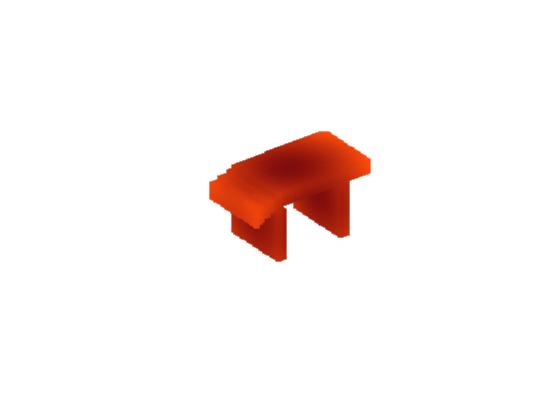} \\      
        \vspace{-0.2cm}
        \includegraphics[width=2cm,trim={4cm 4cm 4cm 4cm},clip]{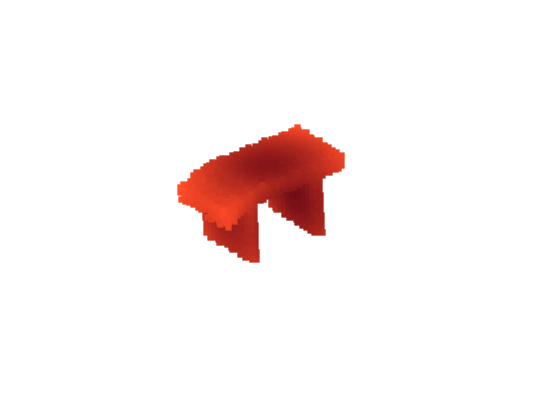}
        \caption{3D-PhysNet}
    \end{subfigure}%
    ~     
    \begin{subfigure}[t]{0.24\columnwidth}
        \centering
        \includegraphics[width=2cm,trim={4cm 4cm 4cm 4cm},clip]{figures/rot/pred2rot.jpg} \\  
         \vspace{-0.2cm}
         \includegraphics[width=2cm,trim={4cm 4cm 4cm 4cm},clip]{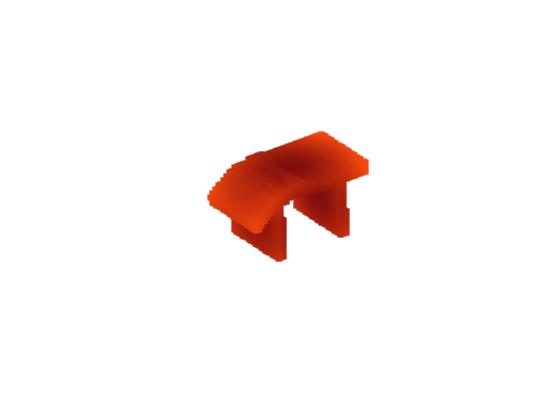}
        \caption{Cascaded-3D-PhysNet}
    \end{subfigure}%
    ~     
    \begin{subfigure}[t]{0.24\columnwidth}
        \centering
        \includegraphics[width=2cm,trim={4cm 4cm 4cm 4cm},clip]{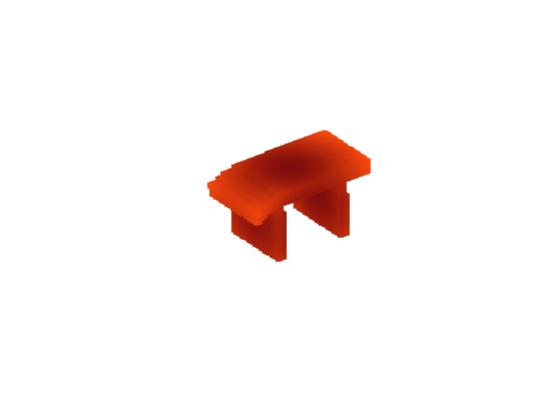} \\
         \vspace{-0.2cm}
        \includegraphics[width=2cm,trim={4cm 4cm 4cm 4cm},clip]{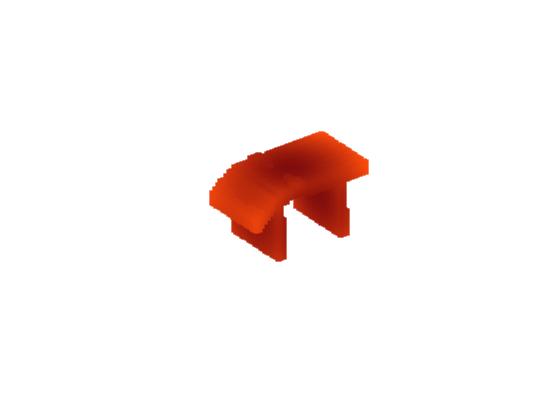}
        \caption{Ground truth}       
    \end{subfigure}%
\caption{Predicted deformations from partial viewpoints.}
\label{fig:partialviewpoints}
\end{figure}

Some works on 3D reconstruction, such as 3D-IwGAN \cite{SmithM17}, leave the object of interest still and rotate the camera around it in order to extract partial 2.5D views. This is a simplified way to extract viewpoints, and our network achieves an accuracy comparable to learning from full 3D objects.
Figure \ref{fig:partialviewpoints} shows some qualitative results.





\subsubsection{Multi-Category and Cross-Category Prediction}
\begin{figure}[t]
\centering
\includegraphics[width=0.75\columnwidth]{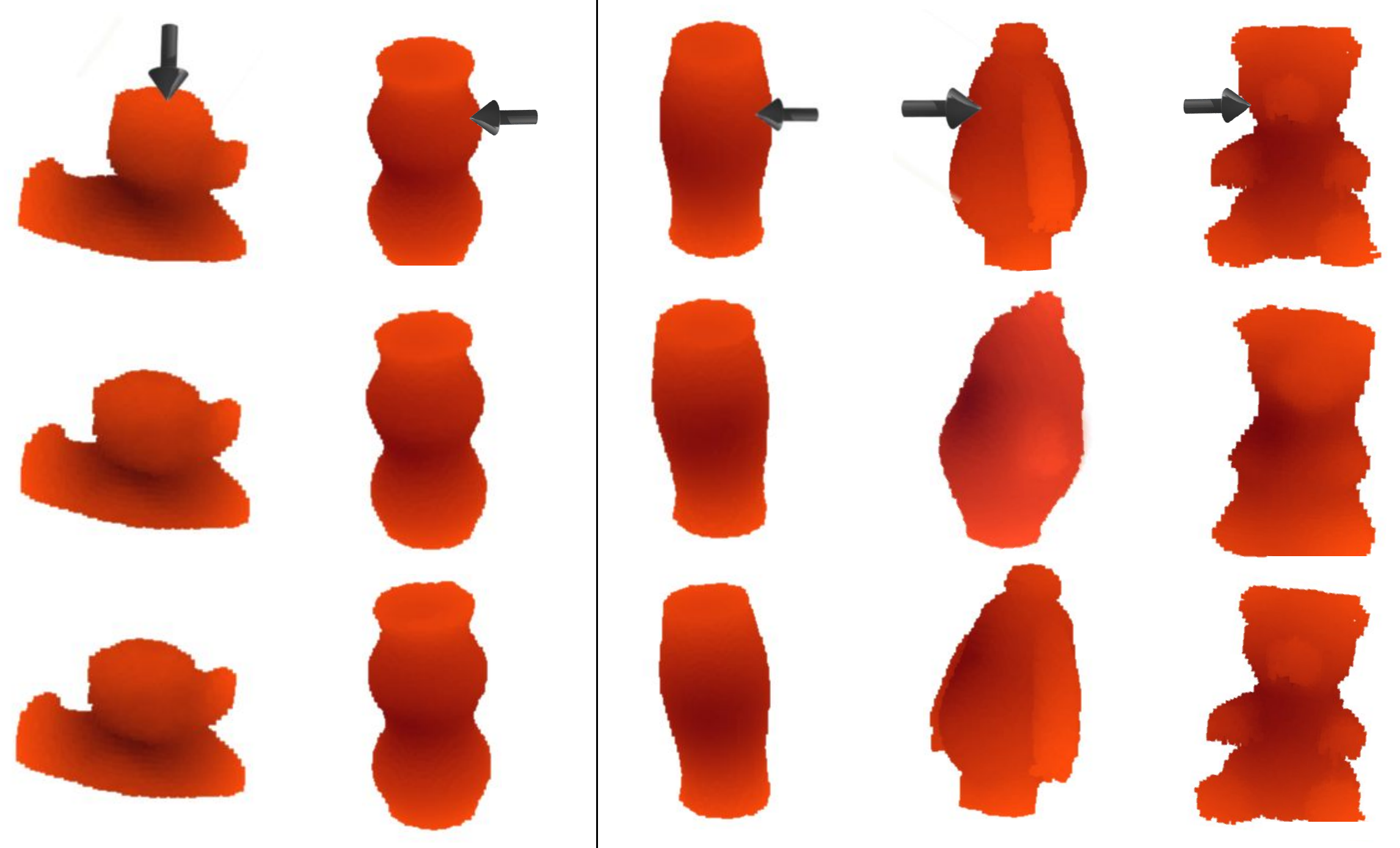}
\caption{Examples of predicted deformations for different objects. Top: undeformed object and applied force; middle: predicted deformation; bottom: ground truth. The three columns on the right were only trained on a set of cylinder-like objects.}
\label{fig:qualitativeresults}
\end{figure}

Finally, we show some qualitative prediction results and cross-category prediction.
The first two columns of Figure \ref{fig:qualitativeresults} shows prediction results on two different objects.
To further investigate the generality of our network we train on a set of cylinder-like objects (second column) and then test on completely unseen objects (last three columns of Figure \ref{fig:qualitativeresults}). 

\begin{figure}[t]
\centering
\includegraphics[width=0.75\columnwidth]{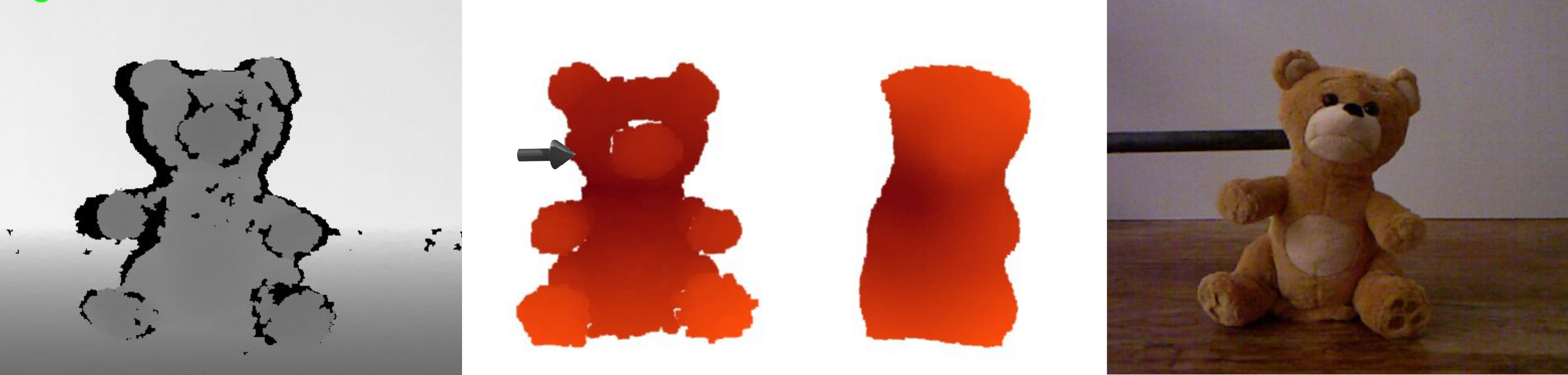}
\caption{Cross-category prediction on a real object. Left: depth input image; middle: input voxel and predicted deformation; right: actual deformation.}
\label{fig:realresults}
\end{figure}

We also show some preliminary results from real depth images in Figure \ref{fig:realresults}.
A single depth view of the real objects is obtained using a Microsoft Kinect camera. 
The results show how the network is able to learn from simple primitives and generalise to unseen objects, both synthetic and from the real world. Note how, albeit simplified in its shape, the predicted shapes of the toys are being pushed by the force (the feet are fixed to the ground) and a simulacrum of the arms is present in the predicted shapes.


\section{Conclusion}
\label{sec:conclusion}
We presented an application of a conditional variational autoencoder architecture with adversarial training to the problem of predicting structural deformations of 3D shapes under the effect of external applied forces, given a single depth image.
The intuition is that the network prediction can be conditioned directly on the elasticity properties of the material, as well as the applied force, enabling the network to learn to approximate non-rigid body deformations of real materials and objects.
This makes the approach useful for a variety of tasks in which it is needed to predict in advance the effect of forces acting on the environment, as diverse as industrial robotics, mobile robotics, structural monitoring, and in general for applications that can benefit from knowledge of intuitive physics.
The network was validated on a large dataset of different objects and conditions.
Future work will be devoted to optimizing the architecture, increasing resolution, and exploring alternative data encodings, such as octrees, surfels, etc.


\bibliographystyle{named}
\bibliography{ijcai18}

\end{document}